\documentclass{article}

\usepackage{microtype}
\usepackage{graphicx}
\usepackage{subfigure}
\usepackage{booktabs} 

\usepackage{hyperref}

\usepackage[accepted]{icml2025}

\usepackage{amsmath}
\usepackage{amssymb}
\usepackage{mathtools}
\usepackage{amsthm}
\usepackage{graphicx}

\usepackage[capitalize,noabbrev]{cleveref}

\theoremstyle{plain}

\theoremstyle{definition}

\theoremstyle{remark}

\usepackage[textsize=tiny]{todonotes}

\begin{document}

\twocolumn[
\icmltitle{Bezier Distillation}



\icmlsetsymbol{equal}{*}

\begin{icmlauthorlist}
\icmlauthor{Ling Feng}{yyy}
\icmlauthor{Sikun Yang}{yyy2}
\end{icmlauthorlist}

\icmlaffiliation{yyy}{Sichuan Agriculture University, Sichuan, China}
\icmlaffiliation{yyy2}{Great Bay University, Guangdong, China}
\icmlcorrespondingauthor{Ling Feng}{202105857@stu.sicau.edu.cn}
\icmlcorrespondingauthor{Firstname2 Lastname2}{first2.last2@www.uk}

\icmlkeywords{Machine Learning, ICML}

\vskip 0.3in
]



\printAffiliationsAndNotice{\icmlEqualContribution} 

\begin{abstract}
In Rectified Flow, by obtaining the rectified flow several times, the mapping relationship between distributions can be distilled into a neural network, and the target distribution can be directly predicted by the straight lines of the flow. However, during the pairing process of the mapping relationship, a large amount of error accumulation will occur, resulting in a decrease in performance after multiple rectifications. In the field of flow models, knowledge distillation of multi - teacher diffusion models is also a problem worthy of discussion in accelerating sampling. I intend to combine multi - teacher knowledge distillation with Bezier curves to solve the problem of error accumulation. Currently, the related paper is being written by myself.

\end{abstract}

\section{Introduction}
One of the main challenges of generative models lies in learning an effective mapping between two distributions. Traditional generative models, such as generative adversarial Networks (GANs\cite{goodfellow2020generative,goodfellow2014generative}) and variational auto-encoders (VAE\cite{kingma2013auto}), attempt to map data points to latent codes that follow a simple base (Gaussian) distribution, through which data can be generated and manipulated. Generative adversarial networks optimize the mapping by introducing a discriminator and utilizing the minimax algorithm, but there are problems such as numerical instability and mode collapse.  Variational auto-encoders introduce the latent variable space and optimize the variational lower bound to approximate the generative distribution, yet they are restricted by their distribution assumptions and reconstruction errors.

While continuous-time methods based on neural ordinary differential equations (ODE) and stochastic differential equations (SDE), provide a new perspective on the mapping problem between two distributions.\cite{chen2018neural,papamakarios2021normalizing,song2020score,ho2020denoising,tzen2019theoretical,de2021diffusion,vargas2021solving}. By taking advantage of the mathematical structures of ODE/SDE, continuous-time models can be trained efficiently without resorting to minimax or traditional approximate inference techniques. For example, score-based generative models\cite{song2019generative,song2020score,song2020improved} and denoising diffusion probabilistic models (DDPMs\cite{ho2020denoising}). Diffusion models utilize stochastic differential equations to model noise diffusion processes and optimize inference speed through probabilistic flow ODE \cite{song2023consistency,song2020score}and denoising diffusion implicit models \cite{songdenoising}. These techniques not only outperform GAN in image generation, but also demonstrate unique advantages in tasks such as domain adaptation, style transfer, audio generation, and video generation\cite{zhu2017unpaired,courty2016optimal,trigila2016data,peyre2019computational,kong2020diffwave,ho2022video,xugeodiff}.They don't have problems of instability and mode collapse\cite{dhariwal2021diffusion,nichol2021glide,saharia2022photorealistic,ramesh2022hierarchical}.

However, continuous-time models have the disadvantage of high computational overhead during the inference stage. For example, ODE/SDE solvers need to call neural networks frequently, and there is no reasonable pairing relationship between noise and data. Moreover, they do not solve the problems of generative modeling and domain transfer. 
The transportation mapping problem is defined as follows:  Given two distributions $\pi_0$ and $\pi_1$ with empirical observations $X_0\sim\pi_0$ and $X_1\sim\pi_1$, the goal is to find a transportation map $T: \mathbb{R}^{d}\to\mathbb{R}^{d}$ , such that for  $X_0\sim\pi_0$ , the resulting $X_1:=T(X_0)$ satisfies $X_1\sim\pi_1$. This problem can be viewed mathematically as the finding of a coupling between the two distributions, which corresponds to the optimal way of redistributing the mass from one distribution to another. \cite{liu2022flow}
To address these issues, recent approaches have proposed transportation ways that optimize the paths \cite{liu2022flow,lipman2022flow,albergo2022building} to reduce computational costs \cite{villani2021topics,ambrosio2021lectures,figalli2021invitation,peyre2019computational}. These models utilize interpolation processes to fit generative ODE models, simplifying the numerical solving process while theoretically ensuring a reduction in transportation costs and the controllability of paths. As a result, they demonstrate high efficiency and robustness in generative modeling and distribution transfer tasks.
Although transportation models that optimize paths provide an efficient continuous-time method, their computational efficiency can still be significantly improved through further distillation. The goal of distillation is to simplify complex multistep transportation models into single-step or few-step models, thereby enabling faster inference. Unlike other knowledge distillation methods \cite{salimans2022progressive,song2023consistency, berthelot2023tract, dockhorn2023distilling, huang2023accelerating}, these approaches introduce additional model training, allowing the student model to learn from the inference samples of the teacher model, effectively reducing the number of steps to a single or few steps. In transportation models that optimize paths, by recursively applying the pairing process of the two distributions, the pairing relationships are distilled into a neural network. The neural network is then utilized to directly approximate the mapping and pairing relationships in transportation models, enabling the input samples to directly generate samples of the target distribution from 0 to 1 through a one-step calculation, without relying on a complete ODE solution. Since the pairing relationships can also be rather complex, if the distillation still attempts to reproduce the pairing relationships between the two distributions in every detail, it will become very difficult to conduct direct distillation\cite{liu2022flow,lipman2022flow,albergo2022building}.

We propose a new method called Bezier distillation, which effectively addresses the distillation challenges caused by complex pairing relationships in transportation models like Rectified Flow\cite{liu2022flow}, by introducing a guiding mechanism. The core idea of transportation models distillation is that, for a given distribution $X_0\sim\pi_0$, the model attempts to directly transfer from $X_0\sim\pi_0$ to the target data distribution $X_1\sim\pi_1$ in a single step during the distillation process. However, we argue that, in the absence of an effective guiding mechanism, this direct path transfer may lead to instability and risks, especially when the pairing relationships are complex.

To address this issue, we introduce one or more intermediate guiding distributions $X_0\sim\pi_0$ located between the initial distribution $X_0\sim\pi_0$ and the target distribution $X_1\sim\pi_1$, relying solely on the initial and target distributions. These guiding distributions are connected through the direction of Bezier curves \cite{Bezier1974mathematical}, forming a smoother and more stable transport path. Algorithmically, by utilizing the guiding mechanism of the intermediate distributions, the model can significantly reduce instability and potential risks during the transport process. On the other hand, due to the inherent smoothness and interpolation properties of Bezier curves at the start and end points, the model can focus on learning the shared features between the guiding distribution and the target distribution $X_1\sim\pi_1$, enabling more efficient transfer to the target distribution and avoiding the limitations that might arise from direct modeling. We implement it on the basis of Rectified Flow\cite{liu2022flow}. In this way, we use the previously obtained Rectified Flow to simulate the reflux process of distilling new Rectified Flow iteratively. We achieve a stable distillation effect in 1-Rectified-Flow and even obtain better performance in 2-Rectified-Flow.
\section{Background}
\subsection{Rectified Flow}

Given the observed data from two empirical distributions $X_0\sim\pi_0$ and $X_1\sim\pi_1$, where $X_0$ is random noise drawn from $\pi_0$ and  $X_1$ is random data drawn from $\pi_1$. Rectified Flow \cite{liu2022flow} is a differential equation (ODE) model defined over the time interval $t\in[0,1]_{:} \frac{dX_{t}}{dt}=v(X_{t},t)$. The Rectified Flow model transforms $X_0$ from distribution $\pi_0$ to $X_1$ so that it follows distribution $\pi_1$. The drift function $v(\mathbb{R}^d\to\mathbb{R}^d)$ is trained to align with the direction of the linear interpolation path from $X_0$ to $X_1$,$i.e., X_t=tX_1+(1-t)X_0$. Thus,$X_t$ satisfies the ODE:$\frac{dX_{t}}{dt}=X_1 - X_0$.To achieve this, Rectified Flow fits the drift function v using a least squares regression problem.
\begin{equation}
    min\int_0^1\mathbb{E}[||(X_1-X_0)-v(X_t,t)||^2]dt,
    \label{eq:minimization}
\end{equation}
where $X_t=tX_1+(1-t)X_0$ is the linear interpolation between $X_0$ and $X_1$. The drift function v is set as a neural network and optimized using stochastic gradient descent or Adam, resulting in our trainable ODE model.

Rectified Flow, expressed in the form of an ODE $\frac{dX_{t}}{dt}=v(X_{t},t)$, ensures the non-intersection of paths, thereby guaranteeing the uniqueness of the solution. This contrasts with linear interpolation paths, which may lead to path crossings. Rectified Flow avoids such crossing phenomena effectively by adjusting the local paths near the crossing points, maintaining distribution consistency while ensuring no intersection. It can be seen as a memoryless particle flow process.

When the objective function is optimized, the pairings $(X_0,X_1)$ generated by Rectified Flow ensure that the transport cost does not increase under all convex cost functions. Unlike randomly independent pairings, the coupling generated by Rectified Flow is deterministic, with a lower transport cost. Its path is nearly a straight line, making numerical simulations more efficient and reducing errors. By recursively applying the Rectified Flow operator, transport costs can be progressively reduced, ultimately achieving an almost perfect linear path. This property significantly lowers the computational cost of continuous-time ODE/SDE models, offering a simplified and efficient simulation approach.

Distillation: By recursively applying the rectification process $X^{k+1}=\text{RectFlow}(\left(X_0^k,X_1^k\right))$ the rectified flow path becomes increasingly straight. After obtaining the $k-th$ rectified flow $X^k$, the relationship between $(X_0^k,X_1^k)$ can be distilled into a neural network to directly predict $X^k$, thereby improving inference speed without the need to simulate the flow. Specifically, if we take $T(X_0 )=X_0+v(X_0)$, the distillation loss function is:
\begin{equation}
    \mathbb{E}[||(X_1^k-X_0^k)-v(X_0^k,0)||^2]dt,
\label{eq:equation2}
\end{equation}
which is a special case of the objective function \eqref{eq:minimization} at t=0. Distillation differs from the rectification process in that distillation aims to faithfully approximate the coupling pair $(X_0,X_1)$, while rectification generates a new coupling pair $(X_0^k,X_1^k)$ with lower transport cost and a straighter flow.

\subsection{Bezier Curve}
The Bezier curve is a smooth curve widely used in fields such as computer graphics, animation, and font design. It is defined by a set of control points and generates points on the curve through the parameter $t\in[0,1]$\cite{Bezier1974mathematical}. The basic form of a Bezier curve is:
\begin{equation}
    \mathbf{B}(t)=\sum_{i=0}^n\binom{n}{i}(1-t)^{n-i}t^i\mathbf{P}_i,
\end{equation}
where $\mathbf{P}_i$ are the control points, and n is the degree of the curve (the number of control points minus 1). 

The Bezier curve is generated progressively through its recursive definition (e.g., the de Casteljau algorithm), which ensures the curve forms a smooth path while maintaining geometric continuity and parametric consistency. The curve's convex hull property guarantees that it remains within the geometric range defined by the control points.

The generation of a Bezier curve follows these basic steps: 1. Perform linear interpolation between all control points $P_0$, $P_1$, … , $P_n$ based on the parameter t. 2. Recursively compute the new control points for each layer until the corresponding point B(t) on the curve is generated\cite{Bezier1974mathematical}. 

The Bezier curve has the following key properties:1. The curve starts at $P_0$ and ends at $P_n$. 2. The curve lies within the convex hull of the control points, ensuring the curve's geometric stability. 3. The shape of the curve is determined by the control points, and the parameter t controls the generation of the curve. With the flexibility of Bezier curves, smooth paths can be constructed, eliminating discontinuities in complex path definitions\cite{Bezier1974mathematical}. 
\begin{figure}
    \centering
    \includegraphics[width=1\linewidth]{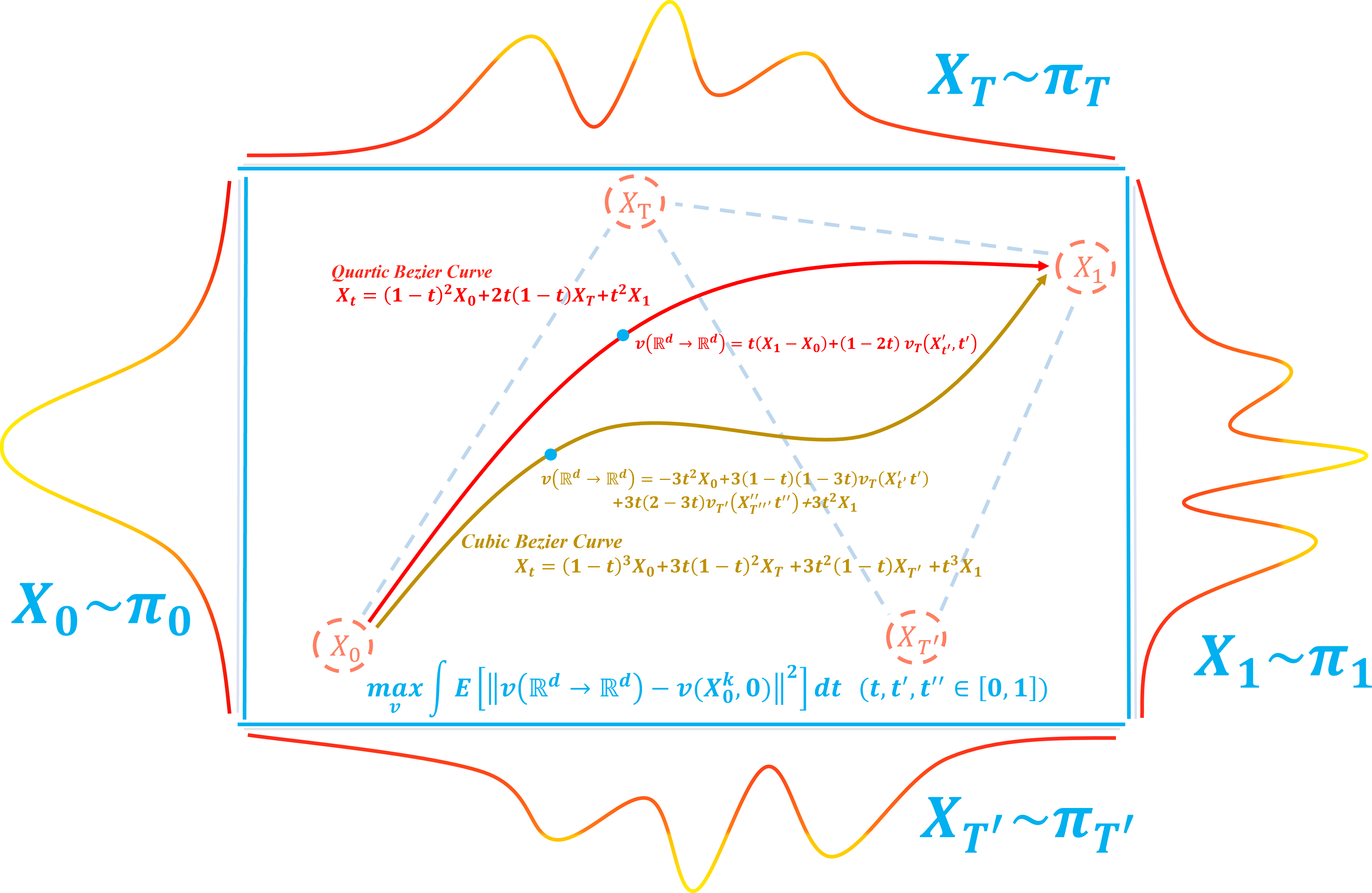}
    \caption{$X_0\sim\pi_1$ and $X_1\sim\pi_1$ are connected by the Bezier curve. Through learning the trajectory, under the guidance of the guiding distribution $X_T\sim\pi_T$, the model is mapped to the target distribution of $X_1\sim\pi_1$ in one step.}
    \label{fig:enter-label}
\end{figure}
\section{Methods}
In flow models, knowledge distillation with multiple teacher diffusion models is also a notable issue for accelerating sampling, which is worth discussing. In CNNs, student models are trained to match the output probability distributions of multiple teachers or to imitate the intermediate layer features of multiple teachers. The final step combines the outputs of the multiple teacher models to construct a composite loss function, such as a trade-off between soft label loss and hard label loss\cite{zhang2018deep,shen2020meal,huang2017snapshot}. However, this approach does not apply directly to diffusion models\cite{de2021diffusion,song2020score,liu2022flow}. In flow models, we use higher-order Bezier curves and the guidance distributions generated by teacher models to effectively guide the initial distribution $X_0\sim\pi_0$ to the target distribution $X_1\sim\pi_1$, thereby achieving knowledge distillation in the multi-teacher diffusion model.

In Rectified Flow, by obtaining the $k$-rectified flow $X_0^k$, the mapping relationship between $(X_0^k,X_1^k)$ can be distilled into a neural network, allowing for the direct prediction of $X_1^k$ and significantly improving inference speed without the need to simulate the flow process step by step. Since the flow is already close to a straight line (and can be well approximated by a single update), this distillation process is highly efficient.

However, the effectiveness of distillation does not improve indefinitely as $k$ increases. This is because, in practical applications, due to the imperfect optimization of $v$, multiple Reflow operations lead to error accumulation. Additionally, although the distillation process of Rectified Flow itself is efficient, performing $k$ iterations of Rectified Flow before distillation is time-consuming.

From the perspective of the objective function's variation, the distillation of Rectified Flow can be seen as a special case of the objective function at t=0. The essence of distillation lies in the model's attempt to directly replicate the process from  $X_0^k$ to $X_1^k$ with a single operation. This effect is entirely dependent on the $k$-Rectified Flow generated $(X_0^k,X_1^k)$ mapping and the model's training performance. Therefore, this method has certain limitations and risks.

Overall, we believe that the problem with Rectified - Flow distillation is that, for a given noise distribution $\pi_0$, the model attempts to directly leap from $\pi_0$ to the target data distribution $\pi_1$ in one step during the distillation process. However, this leap lacks a guiding mechanism. When the pairing relationship is complex, the method has certain limitations and risks. These limitations and risks stem from the inevitable error between  $X_1^k$  generated by k - Rectified Flow and the real data $X_1$, which leads to error accumulation. (See Appendix A for details.) Although Rectified flow can make the particles tend towards $X_1^k$ in a straight - line direction, due to error accumulation, the destination of the particles is not the real data distribution $\pi_1$, and it may even result in larger errors.

On the other hand, apart from the observed data  $X_0\sim\pi_0$ and $X_1\sim\pi_1$ for the given two empirical distributions, as shown in Figure 1, we guide the model to learn from one or more distributions between time 0 and time t (see details below). By utilizing the characteristics of the Bezier curve, we guide the minimization of the objective function from the trajectory, hoping to achieve better results based on Rectified flow.

\subsection{Quartic Bezier Curve}

First, we discuss adding a guiding distribution $X_T\sim\pi_T$ in the transfer from $X_0$ to $X_1$. The guiding distribution $X_T\sim\pi_T$ satisfies that k - Rectified Flow generates  $X_T$  at t=1 from $X_0$ at t=0 in one step, that is,  $X_T=X_0+v_T (X_0,0) $(see the appendix for details). The quadratic Bezier curve\cite{Bezier1974mathematical} is as follows:

First, let's discuss the addition of an intermediate guiding distribution $X_T\sim\pi_T$ when transferring from $X_0\sim\pi_0$ to $X_1\sim\pi_1$. The guiding distribution $X_T\sim\pi_T$ satisfies the condition that, before distillation, the k-Rectified Flow generates a distribution at some time point between t=0 and t=1. During distillation, the model will convert $X_0$ from distribution $\pi_0$ to $X_1$ so that it follows the distribution $\pi_1$. The drift force $v(\mathbb{R}^d\to\mathbb{R}^d)$ is then set to drive the flow in such a way that, under the guidance of distribution $\pi_a$, the flow follows the direction of a quadratic Bezier interpolation path from $X_0$ to $X_1$ :
\begin{equation}
    X_t=(1-t)^2X_0+2t(1-t)X_T+t^2X_1,t\in[0,1],
\end{equation}
where $X_T=X_0+v_T\big(X_0,1).$. 

The curve indicates that starting from time t = 0,$X_0$ is transformed from the $\pi_0$ to $X_1$ under the guidance of the $X_T\sim\pi_0$. That is, knowledge transfer and image generation are achieved. At this time, the drift force $v(\mathbb{R}^d\to\mathbb{R}^d)$ is set to drive the flow as much as possible in the direction of the quadratic Bezier interpolation path from $X_0$ to $X_1$ under the guidance of the distribution $\pi_T$. $X_t$ and the drift force $v(\mathbb{R}^d\to\mathbb{R}^d)$ satisfy the ODE: $\frac{dX_t}{dt}=v(\mathbb{R}^d\to\mathbb{R}^d)=v(X_t,t)=(t-1) X_0+(1-2t) X_T+tX_1$

For the formula [], $X_T=X_0+v_T (X_0,0)$, and after simplification, it can be represented as follows:

Under the quadratic Bezier curve, the drift force $v$ is fitted using a least squares regression problem.
\begin{equation}
\begin{split}
    \min_v\int_0^1\mathbb{E}[||(t-1) X_0+(1-2t) X_T+tX_1-v(X_t,t)||^2]dt,
\end{split}
\label{eq:quartic}
\end{equation}

\begin{equation}
\begin{split}
    \min_v\int_0^1\mathbb{E}[||t(X_1-X_0)+(1-2t)v_T(X_0,1)-v(X_{t},t)||^2]dt,
\end{split}
\label{eq:sim_quartic}
\end{equation}

Where function \ref{eq:sim_quartic} is the simplification of Function \ref{eq:quartic}, and $v(X_t,t)$represents the drift force $v$ at time t. The drift $v$ is overfitted to approximate the objective function.

No matter what order the Bezier curve is, it always connects the initial $X_0\sim\pi_0$ and the target $X_1\sim\pi_1$. The guiding $X_T\sim\pi_T$ only controls the shape of the curve, guiding the original distribution $X_0\sim\pi_0$ toward $X_1\sim\pi_1$, without affecting the final true distribution. Along the points of the distribution, the direction of the curve's tangent is determined by its adjacent control points. Thus, on the Bezier curve, the initial distribution $X_0\sim\pi_0$ will reach the target distribution $X_1\sim\pi_1$ under the guidance of the distribution $X_T\sim\pi_T$. This is different from the initial Rectified Flow distillation, which attempted to rigidly reproduce the paired relationship $(X_0^k,X_1^k)$ without any guidance.

\subsection{Cubic Quartic Bezier Curve/multi teacher}
we discuss adding two guiding distributions $X_T\sim\pi_T$ and $X_{T^{\prime}}\sim\pi_{T^{\prime}}$ in the transmission from $X_0\sim\pi_0$ to $X_1\sim\pi_1$. In this case, the drift force $v(\mathbb{R}^{d}\to\mathbb{R}^{d})$ is set to drive the flow in the direction of the cubic Bezier interpolation path from $X_0$ to $X_1$, guided by the distribution $X_T\sim\pi_T$ and $X_{T^{\prime}}\sim\pi_{T^{\prime}}$.the cubic Bezier interpolation path:
\begin{equation}
\begin{split}
    X_t=(1-t)^3X_0+3t(1-t)^2X_T+3t^2(1-t)X_{T^{\prime}}\\
    +t^3X_1,
\end{split}
\end{equation}
where $X_{T}=X_{0}+v_{T}(X_{t}^{\prime},t^{\prime}),X_{T^{\prime}}=X_{0}+v_{b}(X_{t}^{\prime\prime},t^{\prime\prime}),\\X_{t}^{\prime}=t'X_{1}+(1-t')X_{0},X_{t}^{\prime\prime}=t''X_{1}+(1-t'')X_{0}$.$X_T\sim\pi_T$ and $X_{T^{\prime}}\sim\pi_{T^{\prime}}$ are the guiding distributions at two certain moments between 0 and 1.

Similarly, the drift force $v$ can be fitted using methods such as least squares regression under the cubic Bezier curve:
\begin{equation}
\begin{split}
    \frac{dX_t}{dt}=-3t^2X_0+3(1-t)(1-3t)v_T(X_t^{\prime},t^{\prime})+\\3t(2-3t)v_{T^{\prime}}(X_t^{\prime\prime},t^{\prime\prime})+3t^2X_1,
\end{split}
\end{equation}
\begin{equation}
\begin{split}
    min\int_{v}^{1}\mathbb{E}[||-3t^{2}X_{0}+3(1-t)(1-3t)v_{T}(X_{t}^{\prime},t^{\prime})+\\3t(2-3t)v_{T^{\prime}}(X_{t}^{\prime\prime},t^{\prime\prime})+3t^{2}X_{1}-v(X_{0},0)||^{2}]dt,
\end{split}
\end{equation}
where $v_T(X_t',t' ),v_{T^{\prime}}(X_t'',t'')$ represent the drift forces of the k-Rectified Flow at times $t'$ and $t''$ within the interval [0,1]. $X_{t}'=t' X_1+(1-t')X_0$, $X_t''=t'' X_1+(1-t'' )X_0$.

Regardless of the order of the Bezier curve, its ultimate goal is to guide the initial distribution to the target distribution. This characteristic is determined by the inherent properties of the Bezier curve. Therefore, for the initial distribution and the surrounding guiding distributions, their role is only to guide the initial distribution, without causing the model to truly learn the specific features of these guiding distributions. More precisely, the model is more likely to learn the common features between the guiding distributions and the target distribution, thereby guiding the initial distribution more directionally toward the target distribution.

\subsection{Transport}
Under the effect of the guiding distribution, a causal relationship is presented between the starting point and the ending point. When the starting $X_0$  migrates to $X_1$ , within any time $t\in[1,0]$ , the migration trajectories will not cross each other. That is to say, there does not exist a position $x\in\mathbb{R}^d$ and a time $t\in[1,0]$ such that two paths pass through  along different directions at time  (see the figure).

The Bezier method does not avoid intersections by re - planning each trajectory at the intersection points as in the past. Due to the existence of the guiding distribution, it directly circumvents the intersection situations among the trajectories. In this way, the entire interpolation path can be regarded as a path $X_T$ connecting the relevant points under the guidance of $\pi_T$.

In rectified flow, to obtain accurate rectified flow data pairs $(X_0,X_1)$, the prerequisite is to accurately solve Equation 1 with the help of a numerical solver. Numerical solvers typically discretize the continuous time process into a series of time steps to approximately solve the stochastic differential equation. Within each time step, the stochastic differential equation is approximated. However, this discretization operation will inevitably introduce errors. This is because within each time step, the true solution changes continuously, while the solver can only provide approximate values at discrete time points (see Appendix A for details). Even if there exists  $X_0\sim\pi_0$ and the solved $X_1$ follows $\pi_1$, no matter what numerical solver is used, there will surely be an inevitable error between the obtained  $X1$ and the real data. Therefore, after multiple rectified pairings, the phenomenon of error accumulation is very likely to occur. Although the particle movement paths are straight, the end-points of the paths deviate more and more from the real data distribution, which is caused by the approximate treatment of the numerical solver.

In contrast, although the rectified flow coupling  $(X_0,X_1)$ has a deterministic dependence relationship, the errors generated during the solution process by the numerical solver are still difficult to avoid. We propose a Bezier data pair $(X_0,X_1,X_T)$ . The data pair of the Bezier flow not only has a deterministic dependence relationship but also successfully solves the error problem. The reasons are as follows. On the one hand, in numerical simulations, the flow that is close to the Bezier path uses the guiding distribution to avoid particle crossing, resulting in relatively small time discretization errors. On the other hand, the end - point under the Bezier path is still the real data in the dataset, ensuring the accuracy of path transmission. With the introduction of higher-order Bezier curves, related problems of the multi - teacher diffusion model can also be effectively solved.

\section{Experiment}
Experiment
\section{Conclusion}
We have introduced the Bezier distillation method, which is a better approach for transferring the initial distribution to the target distribution. Through experiments, we have demonstrated that our Bezier distillation method outperforms the current Rectified Flow distillation technique with fewer Rectified Flow iterations. Additionally, the Bezier distillation method generates better samples than existing single-step or two-step generative models or distillation methods. Similar to Rectified Flow, the distilled model also performs well in Image-to-Image Translation tasks.

As the research community continues to explore distillation techniques for flow-based generative models, we believe the Bezier distillation method can provide new directions and insights for the design and optimization of both single-step and multi-step generative models. At the same time, by incorporating more theoretical tools related to distribution transfer, such as optimal transport theory and dynamic programming, we expect this method to show great potential in a variety of multimodal tasks, including image generation, video generation, and text generation.

\section*{Acknowledge}
I'm sorry that I wasn't able to continue finishing the thesis due to my own schedule during the period of visiting students. If you have any ideas, feel free to contact Ling Feng(fengling020928@gmail.com).  
\nocite{langley00}

\bibliography{example_paper}
\bibliographystyle{icml2025}

\newpage
\appendix
\onecolumn
\section{You \emph{can} have an appendix here.}

You can have as much text here as you want. The main body must be at most $8$ pages long.
For the final version, one more page can be added.
If you want, you can use an appendix like this one.  

The $\mathtt{\backslash onecolumn}$ command above can be kept in place if you prefer a one-column appendix, or can be removed if you prefer a two-column appendix.  Apart from this possible change, the style (font size, spacing, margins, page numbering, etc.) should be kept the same as the main body.

\end{document}